\documentclass[runningheads,a4paper]{llncs}

\usepackage{amssymb}
\usepackage{graphicx,epstopdf,slashbox}
\usepackage{subcaption}
\usepackage{url}
\urldef{\mailsa}\path|{alfred.hofmann, ursula.barth, ingrid.haas, frank.holzwarth,|
\urldef{\mailsb}\path|anna.kramer, leonie.kunz, christine.reiss, nicole.sator,|
\urldef{\mailsc}\path|erika.siebert-cole, peter.strasser, lncs}@springer.com|
\newcommand{\keywords}[1]{\par\addvspace\baselineskip
\noindent\keywordname\enspace\ignorespaces#1}

\begin{document}

\mainmatter  

\title{Spontaneous vs. Posed smiles - can we tell the difference?}

\titlerunning{Spontaneous vs. Posed smiles - can we tell the difference?}

%
%
\author{Bappaditya Mandal and Nizar Ouarti}
\institute{Visual Computing Department, Institute for Infocomm Research, Singapore\\ Email address: bmandal@i2r.a-star.edu.sg (Bappaditya Mandal); nizarouarti@gmail.com (Nizar Ouarti)}
\authorrunning{Mandal \emph{et al.}}


%
%

\toctitle{Lecture Notes in Computer Science}
\tocauthor{Authors' Instructions}
\maketitle

\begin{abstract}
Smile is an irrefutable expression that shows the physical state of the mind in both true and deceptive ways. Generally, it shows happy state of the mind, however, `smiles' can be deceptive, for example people can give a smile when they feel happy and sometimes they might also give a smile (in a different way) when they feel pity for others. This work aims to distinguish spontaneous (felt) smile expressions from posed (deliberate) smiles by extracting and analyzing both global (macro) motion of the face and subtle (micro) changes in the facial expression features through both tracking a series of facial fiducial markers as well as using dense optical flow. Specifically the eyes and lips features are captured and used for analysis. It aims to automatically classify all smiles into either `spontaneous' or `posed' categories, by using support vector machines (SVM). Experimental results on large database show promising results as compared to other relevant methods.
\end{abstract}

\section{Introduction}
People believe that human face is the mirror/screen showing internal emotional state of the human body as and when it responds to the external world. This means that, what an individual thinks, feels or understands, etc, deep inside the brain, get imitated into the outside world through its face \cite{Ekman1}. Facial smile expression undeniably plays a huge and pivotal role \cite{Zeng1,Ambadar1,Hoque1} in understanding social interactions within a community. People often give smile imitating the internal state of the body. For example, generally, people smile when they are happy or when sudden humorous things happen/appear in front of them. However, people are sometimes forced to pose smile because of the outside pressure or external factors. For example, people would pose a smile even when they don't understand the joke or the humor. Sometimes people would also pose a smile even when they are reluctantly or unwillingly do or perform something in front of their bosses/peers \cite{Ekman3}.

Therefore being able to identify the type of smiles of individuals would give affective computing a deeper understanding of the human interactions. A large amount of research in psychology and neuroscience studying facial behavior demonstrate that spontaneous deliberately displayed facial behavior has differences both in utilized facial muscles and their dynamics as compared to posed ones \cite{Ekman5}. For example, spontaneous smiles ar smaller in amplitude, longer in duration, slower in onset and offset times than posed smiles \cite{Cohn1,Ekman5,Valstar1}. For humans, capturing such subtle facial movements is difficult and we often fail to distinguish between them. It is not surprising that in computer vision, algorithms developed for classifying such pose and spontaneous smiles usually fail to generalize to the subtlety and complexity of human pose and spontaneous affective behavior \cite{Zeng1}.

Numerous researchers asserted that dynamic features such as duration and speed of the smile play a part in differentiating the nature of the smile \cite{Hoque1}. A spontaneous smile usually take longer time to reach from onset to apex and then offset as compared to a posed smile \cite{Dibeklioglu1}. As for non-dynamic features, the aperture size of the eyes is found to be a useful clue and is generally of a higher value when extracted from a spontaneous smile as compared to a posed one. On the other hand, the symmetry in (or the lack of) movement of spontaneous and posed smiles do not produce significant distinction in identifying them and is therefore not much useful \cite{Schmidt1}. In \cite{Valstar1} a multi-modal system using geometric features such as shoulder, head and inner facial movements are fused together and GentleSVM-sigmod is used to classify the posed and spontaneous smiles. He \textit{et al.} in \cite{He3} proposed a technique for feature extraction and compared the performance using geometric and facial appearance features. Appearance based features are computed by recording statistics of overall pixel values of the image, or even using edge detection algorithm such as Gabor Wavelet Filter. Their comprehensive study shows that geometric features are generally more effective in detecting posed from spontaneous expressions \cite{He3}.

A spatiotemporal method involving both natural and infrared face videos to distinguish posed and spontaneous expressions is proposed in \cite{Pfister1}. Using temporal space and image sequences as volume, they extended the complete local binary patterns texture based descriptor into the spatiotemporal features to classify posed and spontaneous smiles. Dibeklioglu \textit{et al.} in \cite{Dibeklioglu2} used the dynamics of eyelid movements and defined distance based and angular features in the changes of the eye aperture. Using several classifiers they have shown the superiority of eyelid movements over the eyebrows, cheek and lip movements for smile classification.
Later in \cite{Dibeklioglu1}, they used dynamic characteristics of eyelid, cheek and lip corner movements for classifying posed and spontaneous smiles. Temporal facial information is obtained in \cite{Huijser1} through segmenting the facial expression into onset, apex and offset which cover the entire duration of the smile. They reported good classification performance on a small database by using a combination of features extracted from the different phases.

\begin{figure}[!ht]
\begin{center}
\scalebox{0.45}{\rotatebox{0}{\includegraphics*{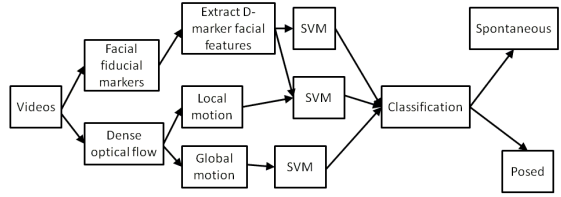}}}
\caption{Block diagram of the proposed system.} \label{BlockDiagram}
\end{center}
\end{figure}

The block diagram of our proposed method is shown in Fig. \ref{BlockDiagram}. Given smile video sequences of various subjects, we apply the facial features detection and tracking of the fiducial points over the entire smile video clip. Using D-markers, 25 important parameters (like duration, amplitude, speed acceleration, etc) are extracted from two important regions of the face: eyes and lips. Smile discriminative features are extracted using dense optical flow along the temporal domain from the global (macro) motion and local (micro) motion of the face. All these information are fused and support vector machine (SVM) is then used as a classifier on these parameters to distinguish posed and spontaneous smiles. \vspace{-0.5cm}

\vspace{-1.25cm}

\section{Feature Extraction from Various Face Components} \vspace{-0.1cm}
We use the facial tracking algorithm developed by Nguyen \textit{et al.} in \cite{Nguyen1} to obtain the fiducial points on the face. The 21 tracking markers each are labeled and placed following the convention as shown in Fig. \ref{FacialMarkersandTracking} (a). The markers are manually annotated in the first frame of each video by user input and thereafter it automatically tracks the remaining frames of the smile video, it is of good accuracy and precision as compared to other facial tracking software \cite{Asthana1}. The markers are placed on important facial feature points such as eyelids and corner of the lips for each subject. The convention followed in our approach for selecting fiducial markers are shown in Fig. \ref{FacialMarkersandTracking} (a). \vspace{-0.2cm}

\subsection{Face Normalization}
\begin{figure}
    \centering
    \begin{subfigure}[b]{0.23\textwidth}
        \centering
        \includegraphics*[height=2.65cm]{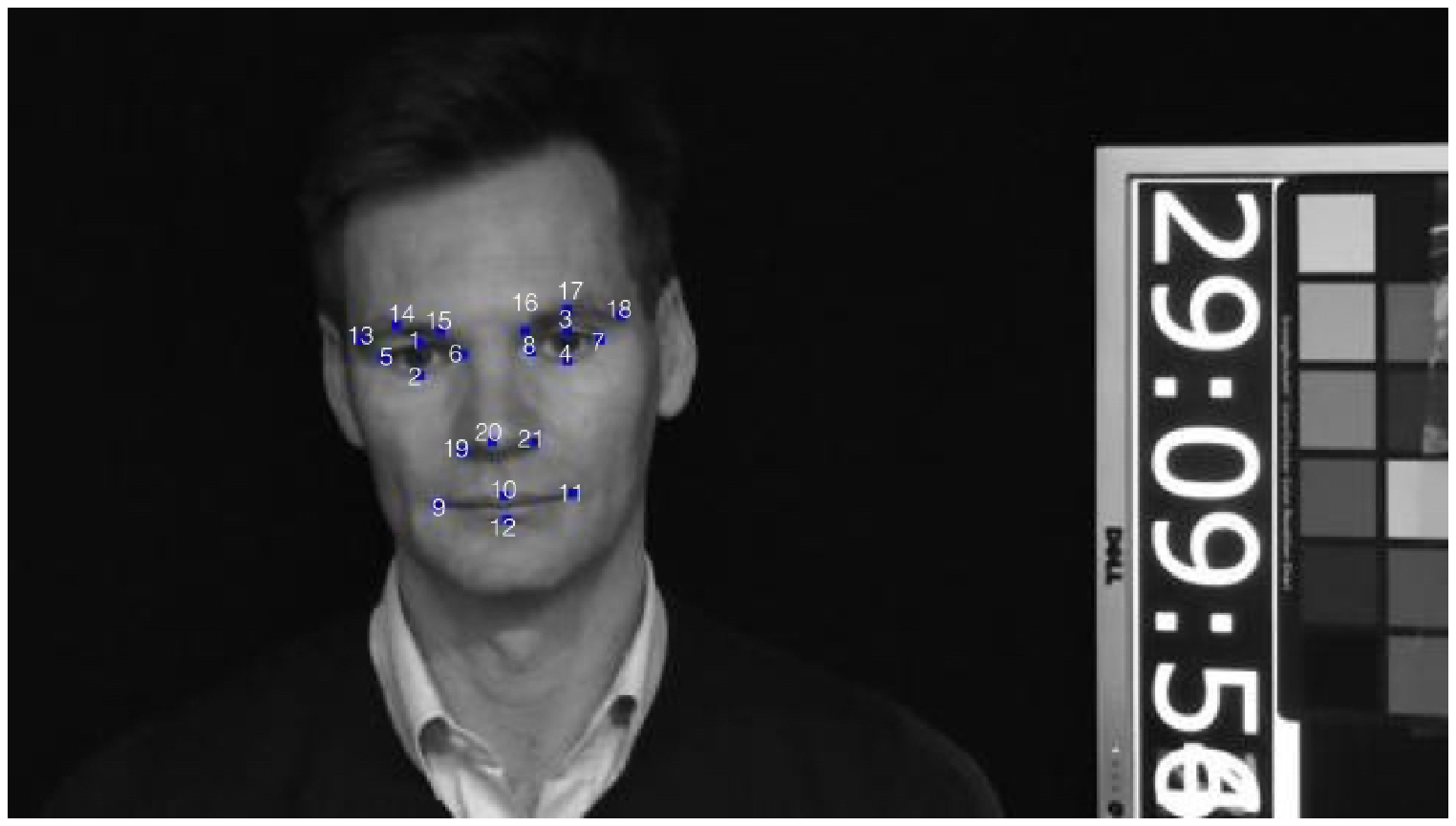} \vspace{-0.75cm}
        \caption{Frame \#1}
        \label{a}
    \end{subfigure}
    \begin{subfigure}[b]{0.23\textwidth}
        \centering
        \includegraphics*[height=2.65cm]{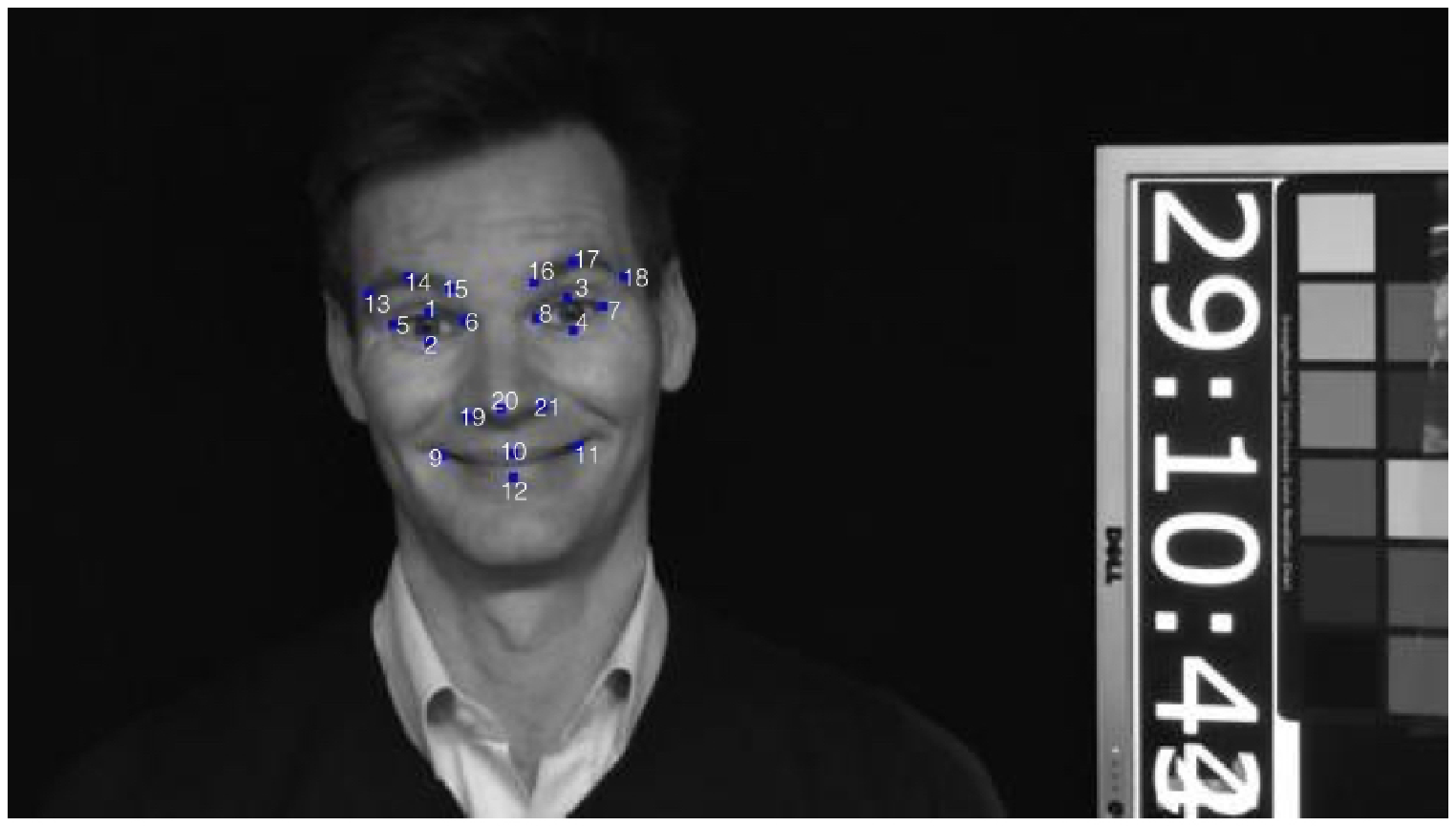} \vspace{-0.75cm}
        \caption{Frame \#30}
        \label{b}
    \end{subfigure}
    \begin{subfigure}[b]{0.23\textwidth}
        \centering
        \includegraphics*[height=2.65cm]{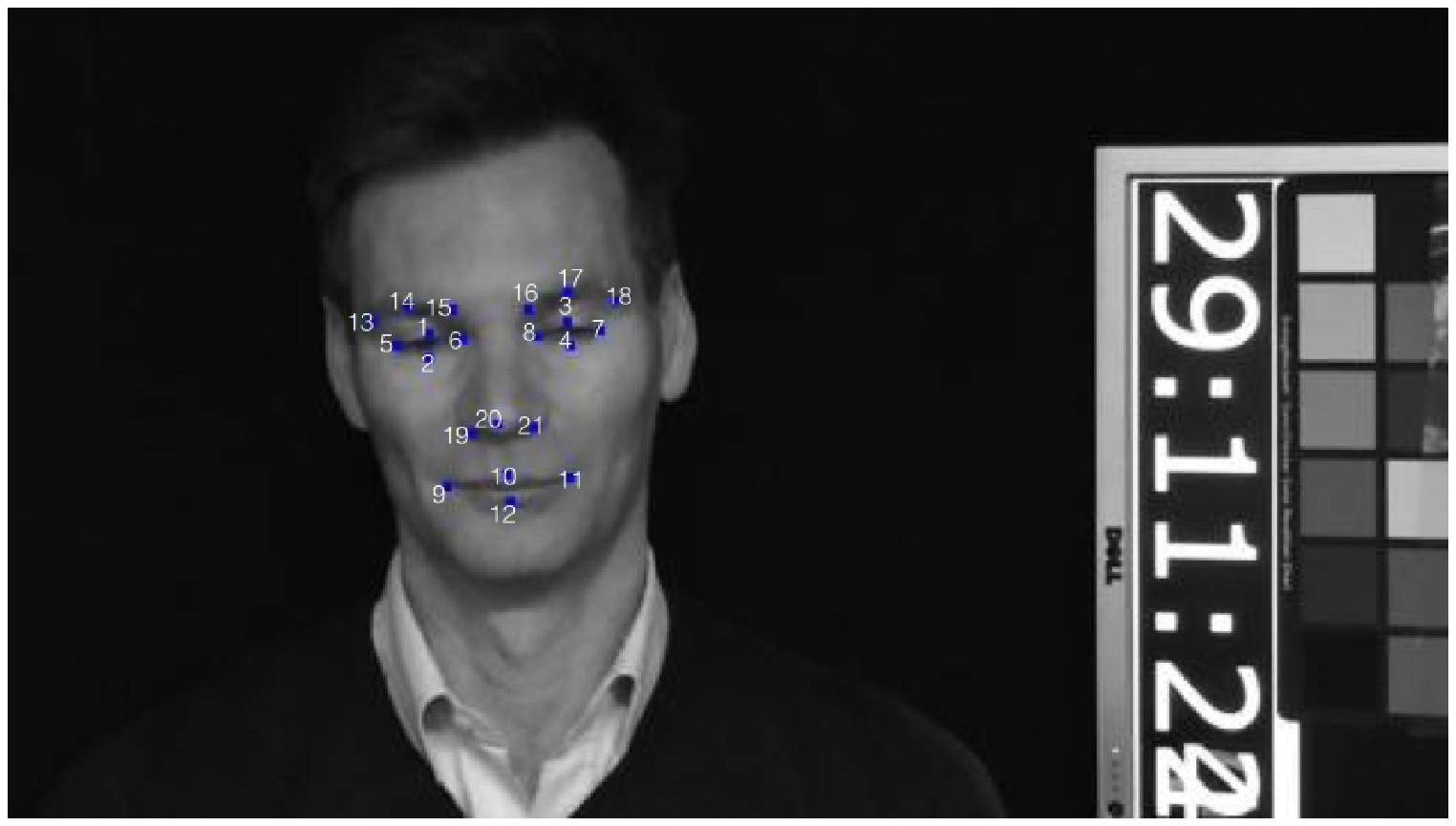} \vspace{-0.75cm}
        \caption{Frame \#58}
        \label{c}
    \end{subfigure}
    \begin{subfigure}[b]{0.23\textwidth}
        \centering
        \includegraphics*[height=2.65cm]{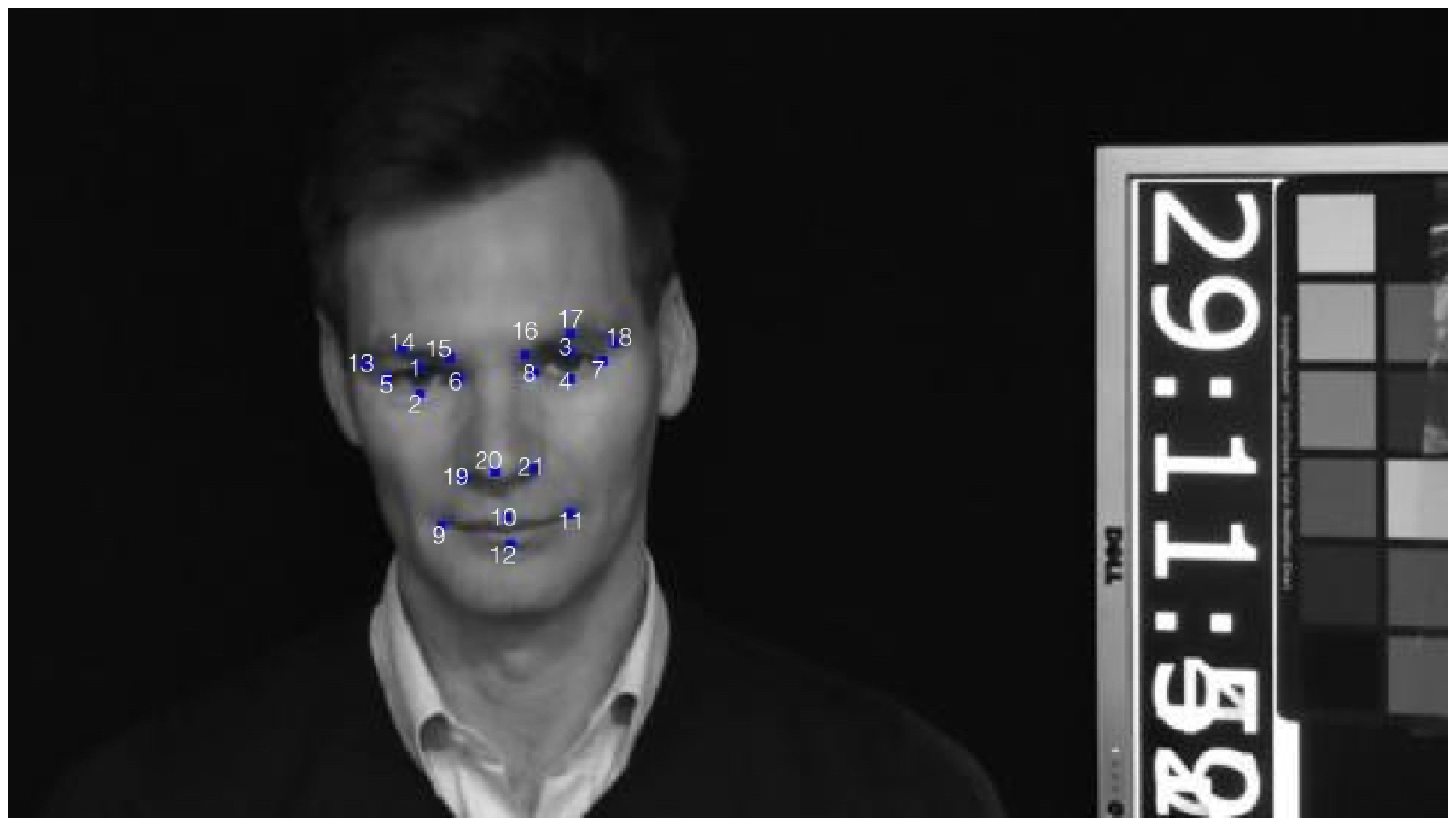} \vspace{-0.75cm}
        \caption{Frame \#72}
        \label{d}
    \end{subfigure}
    \caption{(a) Shows the tracked points on the $1^{st}$ frame, (b) shows the tracked points on $30^{th}$ frame, (c) shows the tracked points on $58^{th}$ frame and (d) shows the tracked points on $72^{nd}$ frame on one subject. (Best viewed when zoomed in.)}
    \label{FacialMarkersandTracking}
\end{figure}

To reduce inaccuracy due to the subject's head motion in the video that can cause change in angle with respect to roll, yaw and pitch rotations, we use the face normalization procedure described in \cite{Dibeklioglu1}. Let $l_i$ represents each of the feature points used to align the faces as shown in Fig. \ref{FacialMarkersandTracking}. Three non-collinear points (eye centers and nose tip) are used to form a plane $\rho$. Eye centers are defined as $c_1=\frac{l_1+l_3}{2}$ and $c_2=\frac{l_4+l_6}{2}$. Angles between the positive normal vector $N_\rho$ of $\rho$ and unit vectors \textit{U} on \textit{X} (horizontal), \textit{Y} (vertical), and \textit{Z} (perpendicular) axes give the relative head pose as follows:
\begin{equation}\label{eq:1}
\theta = arccos \frac{U.N_\rho}{\|U\| \|N_\rho\|}, \textmd{where} ~ N=\overrightarrow{l_gc_2} \times \overrightarrow{l_gc_1}.
\end{equation}
$\overrightarrow{l_gc_2}$ and $\overrightarrow{l_gc_1}$ denote the vectors from point $l_g$ to points $c_2$ and $c_1$, respectively. $\|U\|$ and $\|N_\rho\|$ represents the magnitudes of $U$ and $N_\rho$ vectors respectively.
Using the human face configuration, (\ref{eq:1}) can estimate the exact roll ($\theta_z$) and yaw ($\theta_y$) angles of the face with respect to the camera. If we start with the frontal face, the pitch angles ($\theta'_x$) can be computed by subtracting the initial value. Using the estimated head pose, tracked fiducial points are normalized with
respect to rotation, scale and translation as follows:
\begin{equation}\label{eq:2}
l'_i=[l_i-\frac{c_1+c_2}{2}] R_x(-\theta'_x) R_y(-\theta_y) R_z(-\theta_z) \frac{100}{\epsilon(c_1+c_2)},
\end{equation}
where $l'_i$ is the aligned point. $R_x$, $R_y$ and $R_z$ denote the 3D rotation matrices for the given angles. $\epsilon()$ is the Euclidean distance measure. Essentially (\ref{eq:1}) constructs a normal vector perpendicular to the plane of the face using three points (nose tip and eye centers), then calculate the angle formed between $X$, $Y$ and $Z$ axis with regards to the normal vector of face plane. Thereafter, (\ref{eq:2}) process and normalize each and every point of the frame accordingly and set the interocular distance to 100 pixels with the middle point acting as the new origin of the face center. \vspace{-0.75cm}

\vspace{-1cm}

\subsection{D-Marker Facial Features}
In the first part of our strategy, we focus on extracting the subject's eyelid and lips features. We first construct a amplitude signal variable based on the facial feature markers on the eyelid regions. We compute the amplitude of eyelid and lip end movements during a smile using the procedure described in \cite{Schmidt1}. Eyelid amplitude signals are computed using the eyelid aperture ($D_{eyelid}$) displacement at time $t$, given by: \vspace{-0.3cm}
\begin{equation}\label{eq:3}
D_{eyelid}(t)=\frac{\kappa(\frac{l^t_1+l^t_3}{2}, l^t_2) \epsilon(\frac{l^t_1+l^t_3}{2}, l^t_2)+\kappa(\frac{l^t_4+l^t_6}{2}, l^t_5) \epsilon(\frac{l^t_4+l^t_6}{2}, l^t_5)}{2\epsilon(l^t_1,l^t_3)}
\end{equation}
where $\kappa(l_i, l_j)$ denotes the relative vertical location function, which equals to -1 if $l_j$ is located (vertically) below $l_i$ on the face, and 1 otherwise. The equation above uses the markers for eyelids namely 1-6 as shown in Fig. \ref{FacialMarkersandTracking}, to construct the amplitude signal that calculate the eyelid aperture size in each frame $t$. The amplitude signal $D_{eyelid}$ is then further computed to obtain a series of features. In addition to the amplitudes, speed and acceleration signal are also extracted by computing the second derivatives of the amplitudes.

Smile amplitude is estimated as the mean amplitude of right and left lip corners, normalized by the length of the lip. Let $D_{lip}$(t) be the value of the mean amplitude signal of the lip corners in the frame $t$. It is estimated as
\begin{equation}\label{eq:4}
D_{lip}(t)=\frac{\epsilon(\frac{l^t_{10}+l^t_{11}}{2}, l^t_{10}) + \epsilon(\frac{l^t_{10}+l^t_{11}}{2}, l^t_{11})}{2\epsilon(l^t_{10},l^t_{11})}
\end{equation}
where $l^t_i$ denotes the 2D location of the $i^{th}$ point in frame $t$. For each video of our subject we are able to acquire a 25-dimensional feature vectors based on the eyelids markers and lip corner points. Onset phase is defined as the longest continuous increase in $D_{lip}$. Similarly, the offset phase is detected as the longest continuous decrease in $D_{lip}$. Apex is defined as the phase between the last frame of the onset and the first frame of the offset. The displacement signals of eyelids and lip corners could then be calculated using the tracked points. Onset, apex and offset phases of the smile are estimated using the maximum continuous increase and decrease of the mean displacement of the eyelids and lip corners. The D-Marker is then able to extract 25 descriptive features each for eyelids and lip corner, so a vector of 50 features are obtained from each frame (using two frames at a time). The features are then concatenated and passed through SVM for training and classification.

\subsection{Features from Dense Optical Flow}
\label{SectionDenseOpticalFlow}
\begin{figure}
\centering
\begin{minipage}[b]{14.3cm}
\includegraphics*[height=2.0cm]{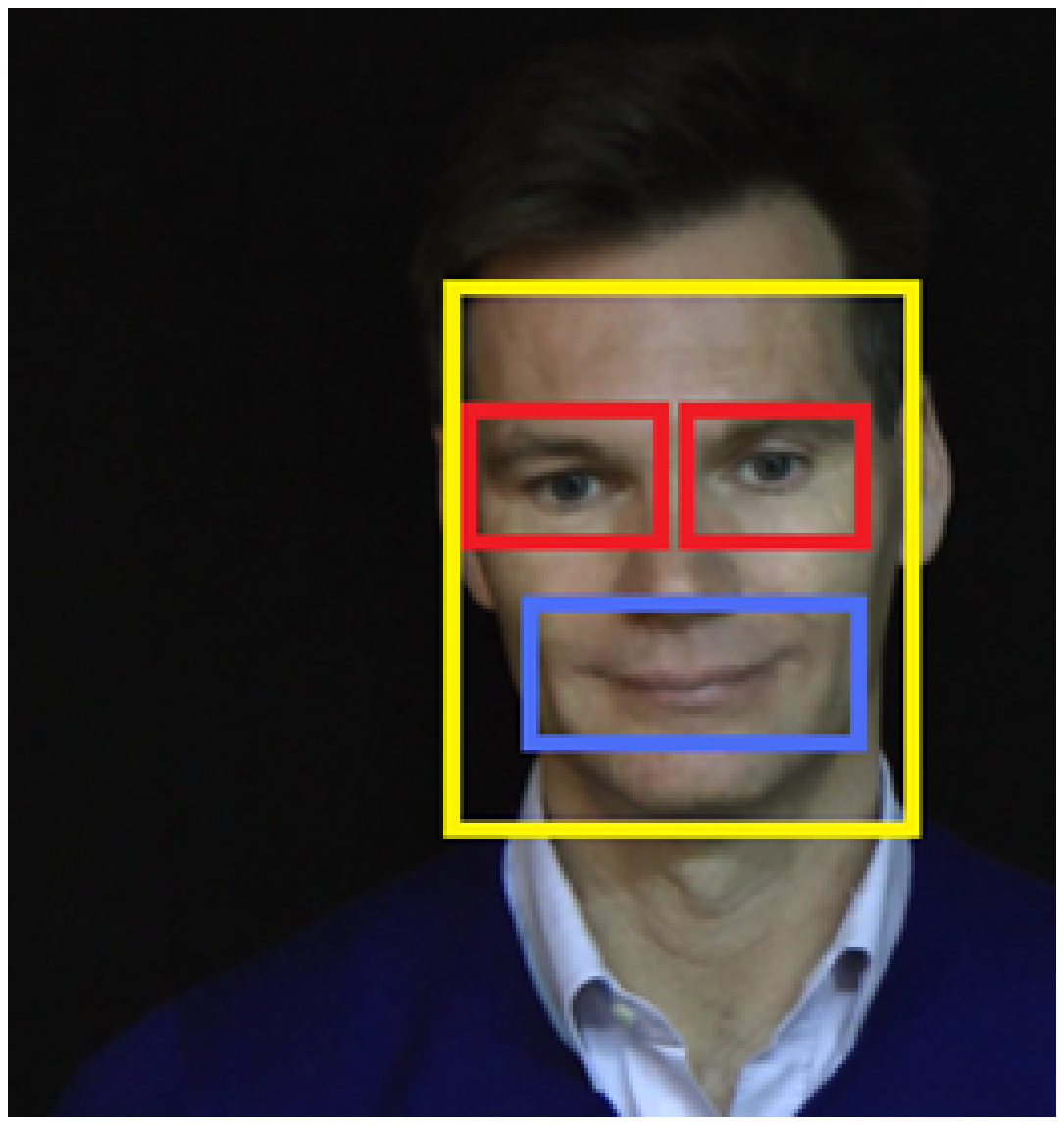}
\includegraphics*[height=2.0cm]{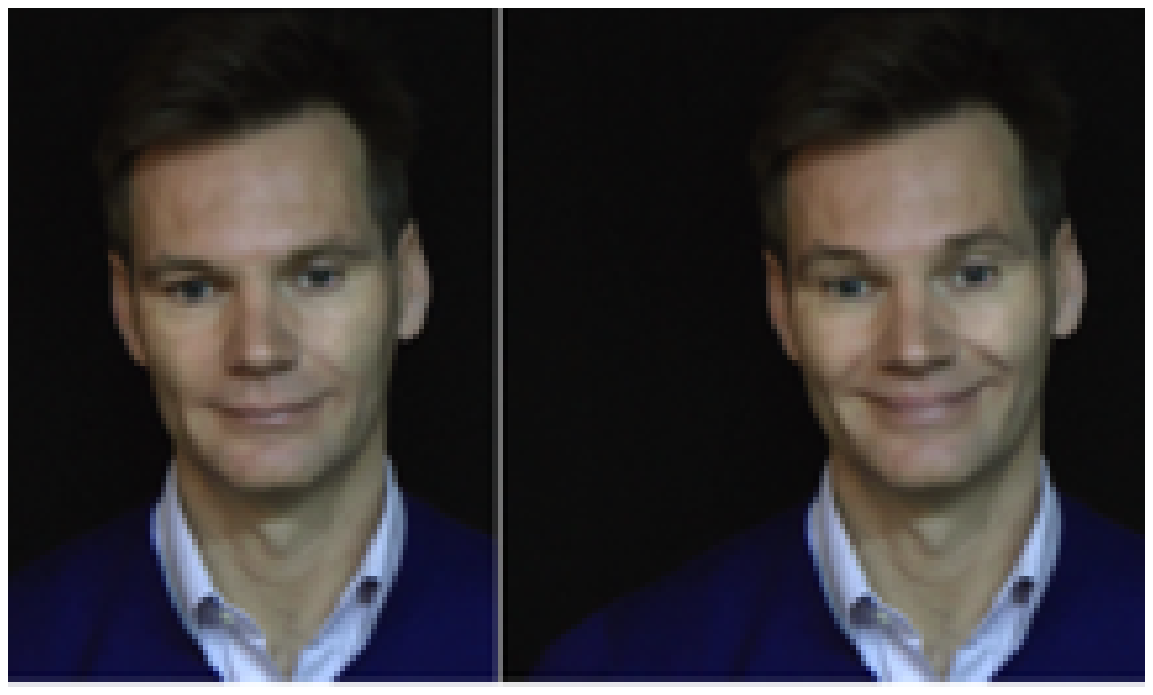}
\includegraphics*[height=2.0cm]{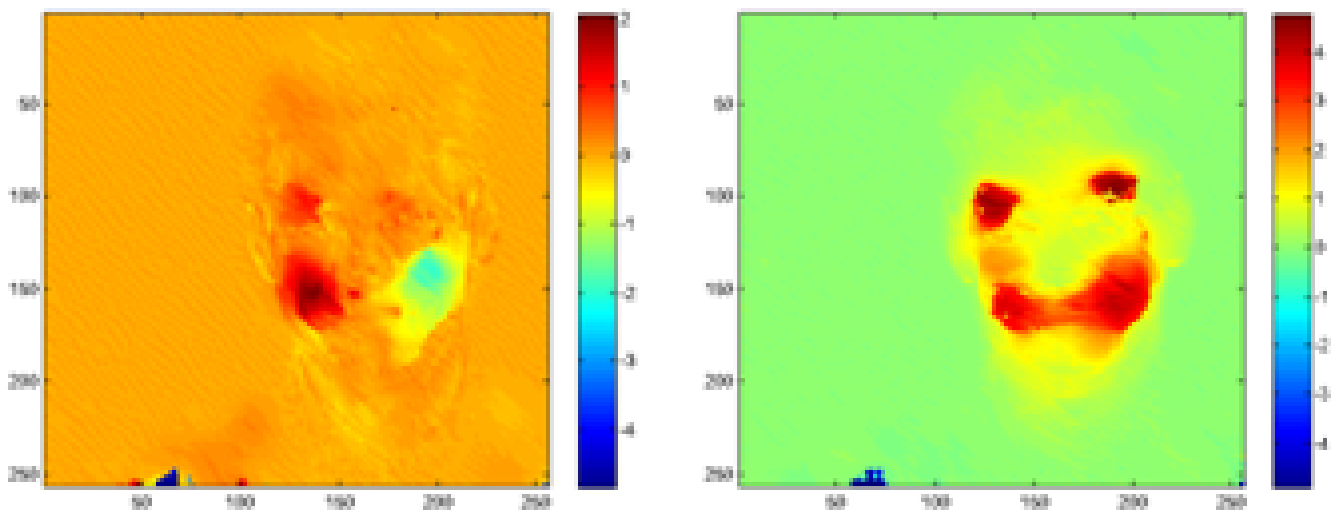}
\end{minipage}
\caption{Left: Face, eyes and mouth detections. Yellow ROI for face detection, red ROI for eyes detection and blue ROI for mouth detection. Middle: Two consecutive frames of a subject's smile video and Right: their optical flows in x- and y-directions. (Best viewed in color and zoomed in.)}
\label{ImageandOpticalFlow}
\end{figure}

In the second phase of the feature extraction, we use our own proposed dense optical flow \cite{ouarti2013method} for capturing both global and local motions appearing in the smile videos. Our approach is divided into four distinct stages that are fully automatic and does not require any human intervention. The first step is to detect each frame in which the face is present. We use our previously developed face, integration of sketch and graph patterns (ISG) eyes and mouth detectors for face recognition on wearable devices and human-robot-interaction \cite{Mandal5,Yu2,Mandal2}. So we get the region of interest (ROI) for the face (as shown in Fig. \ref{ImageandOpticalFlow}, top left, yellow ROI) with 100\% accuracy on the entire UvA-NEMO smile database \cite{Dibeklioglu1}. In the second step, we determine the area corresponding to the right eye, left eye in red ROI and mouth in blue ROI (see Fig. \ref{ImageandOpticalFlow} top left) for which we get 96.9\% accuracy on the entire database. \vspace{-0.5cm}

\vspace{-0.5cm}

\begin{figure}[!htp]
\centering
\begin{minipage}[b]{13cm}
\includegraphics*[height=1.8cm]{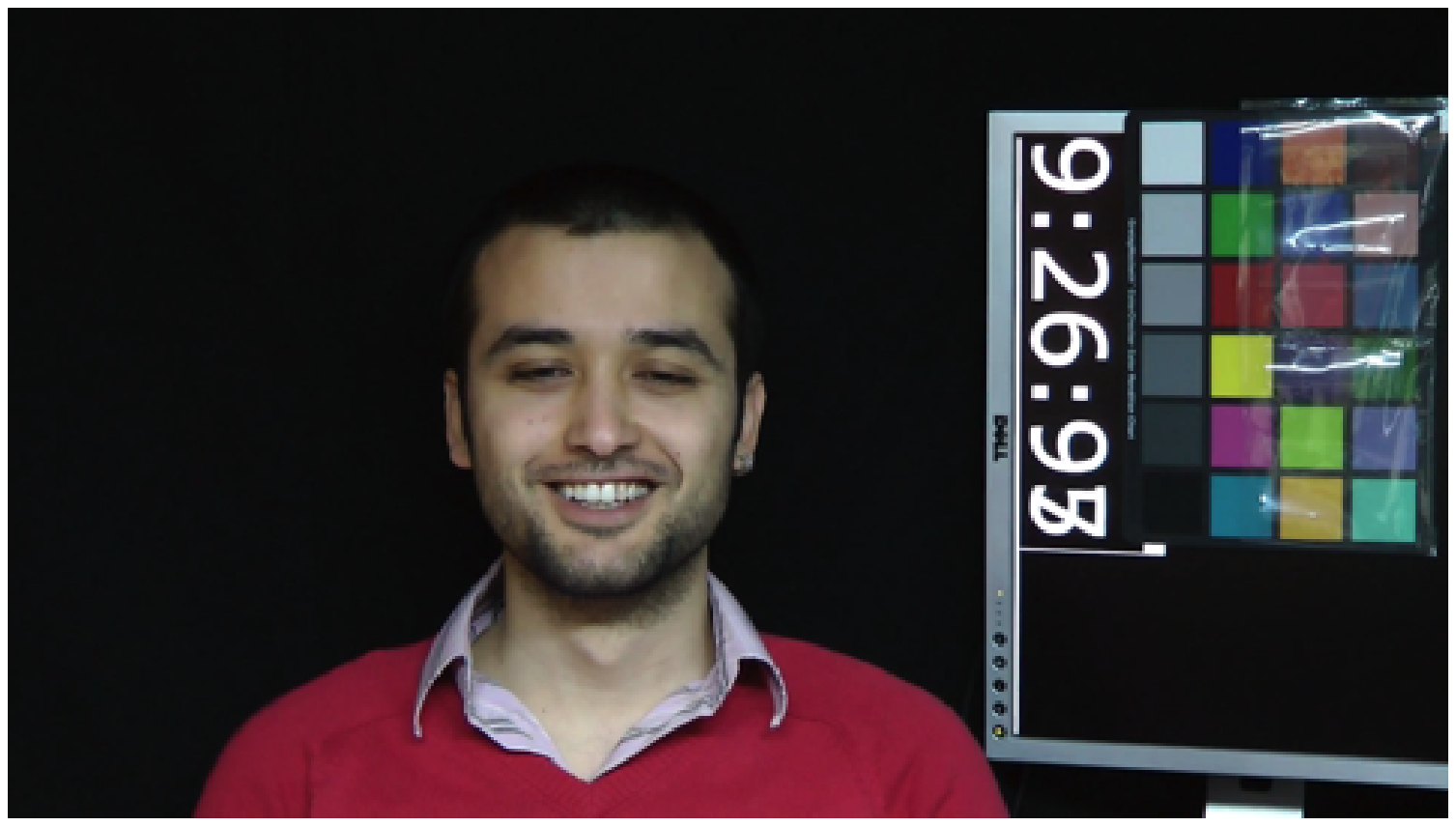}
\includegraphics*[height=1.8cm]{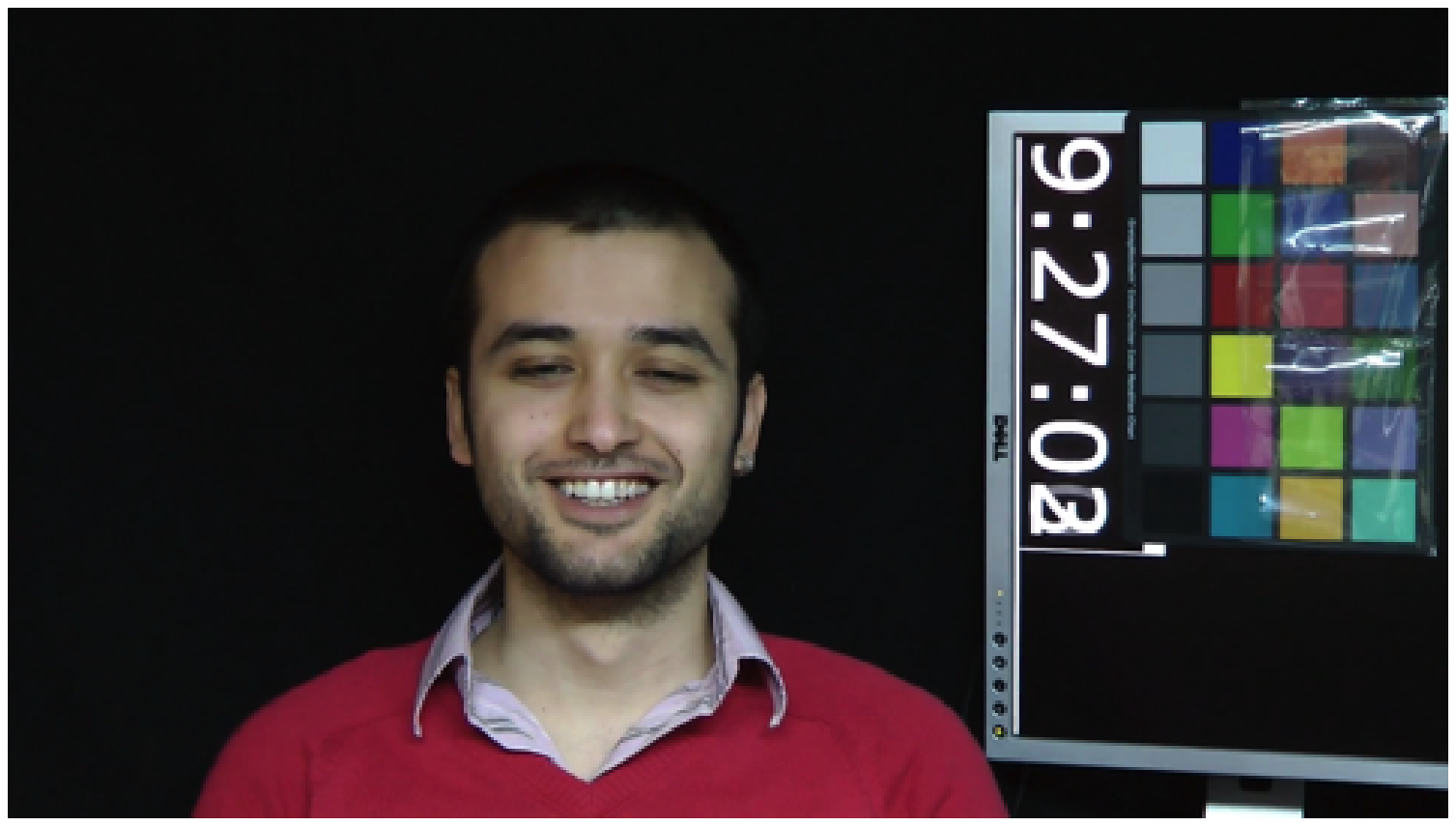}
\includegraphics*[height=1.8cm]{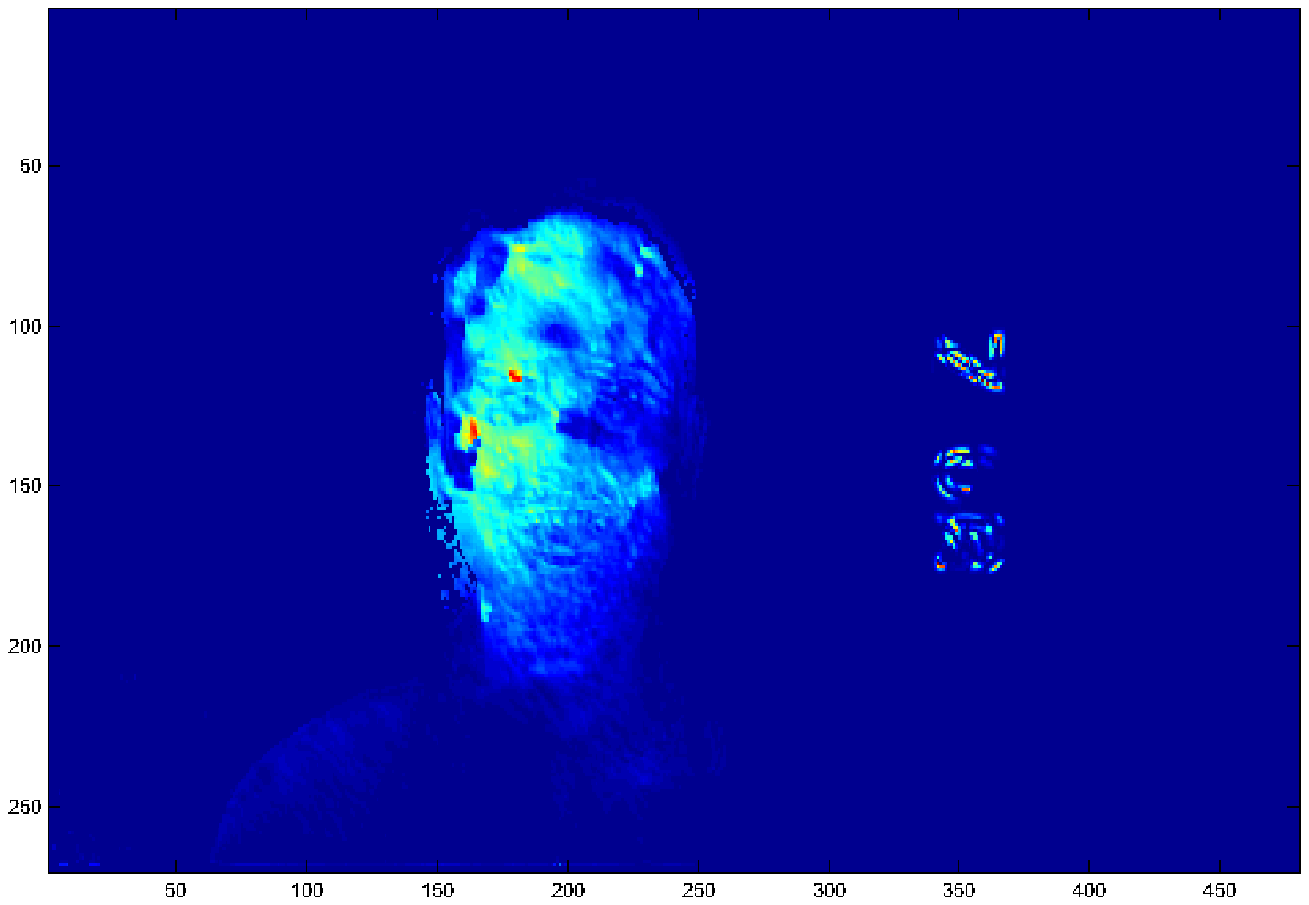}
\includegraphics*[height=1.8cm]{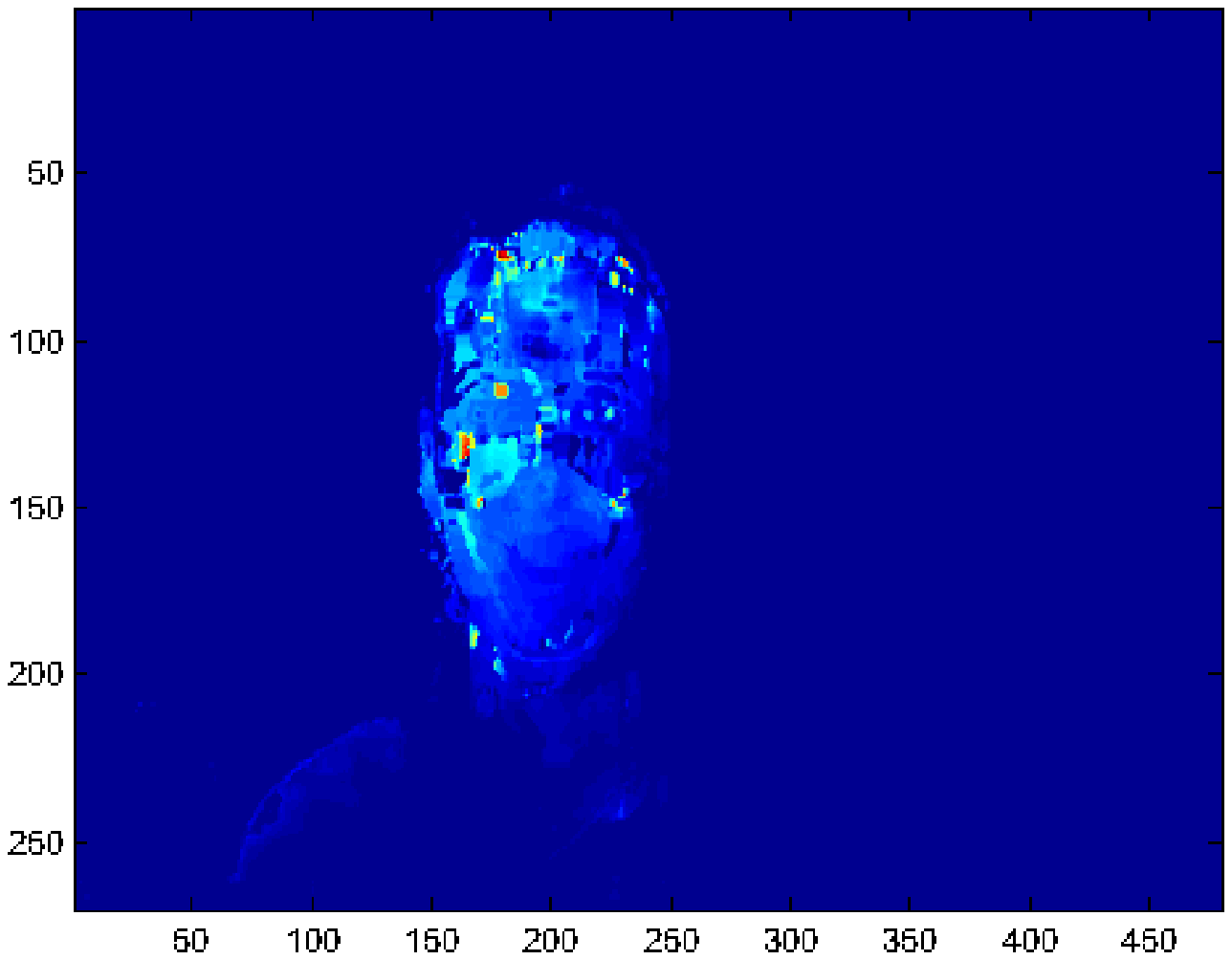}
\end{minipage}
\caption{Original images and their dense optical flows with their corresponding micro and macro motion variations of a subject. (Best viewed in color and zoomed in.)} \label{fig3_leftEandrightE}
\end{figure}

\begin{figure}[!htp]
\centering
\begin{minipage}[b]{13cm}
\includegraphics*[height=1.8cm]{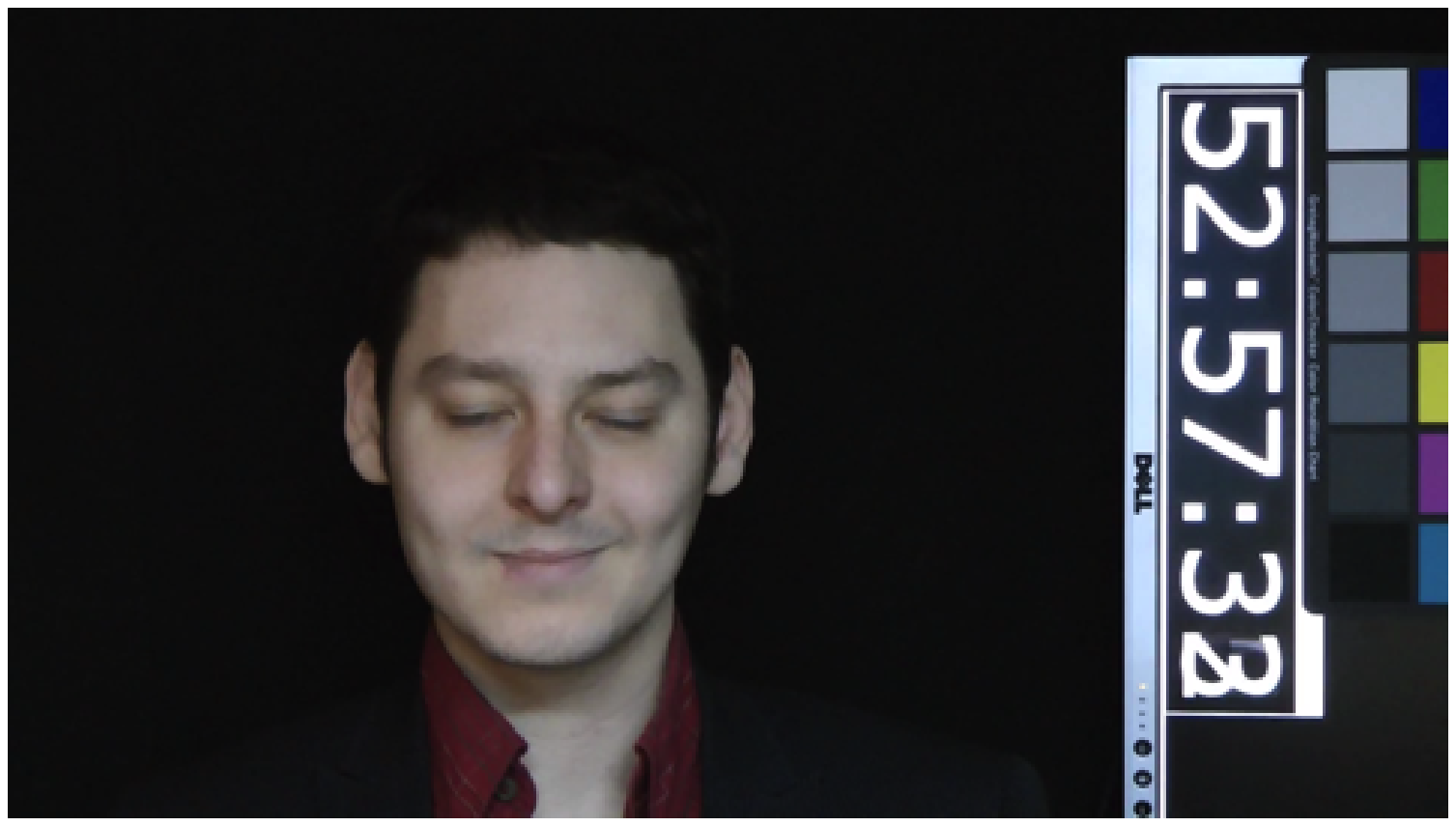}
\includegraphics*[height=1.8cm]{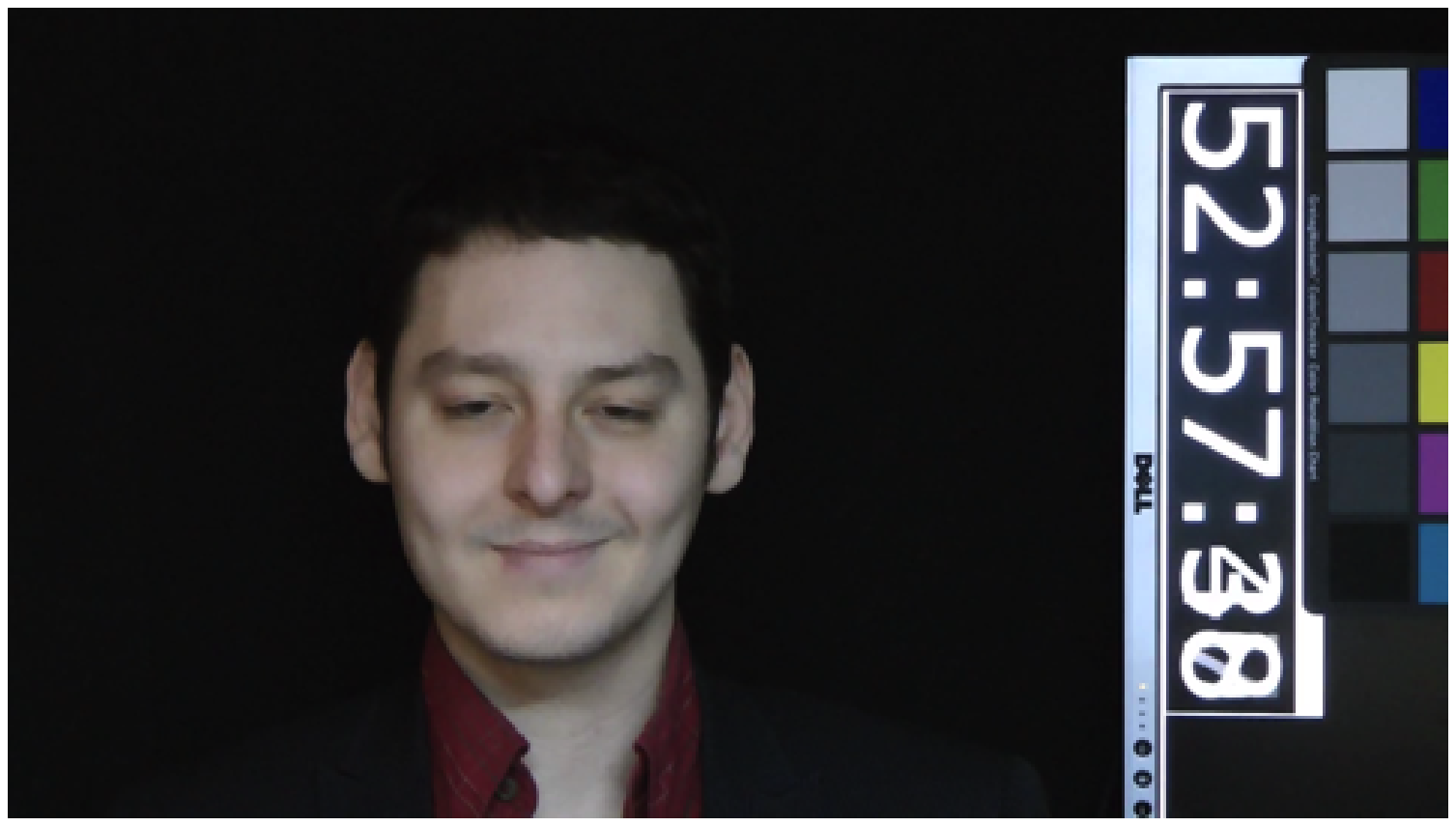}
\includegraphics*[height=1.8cm]{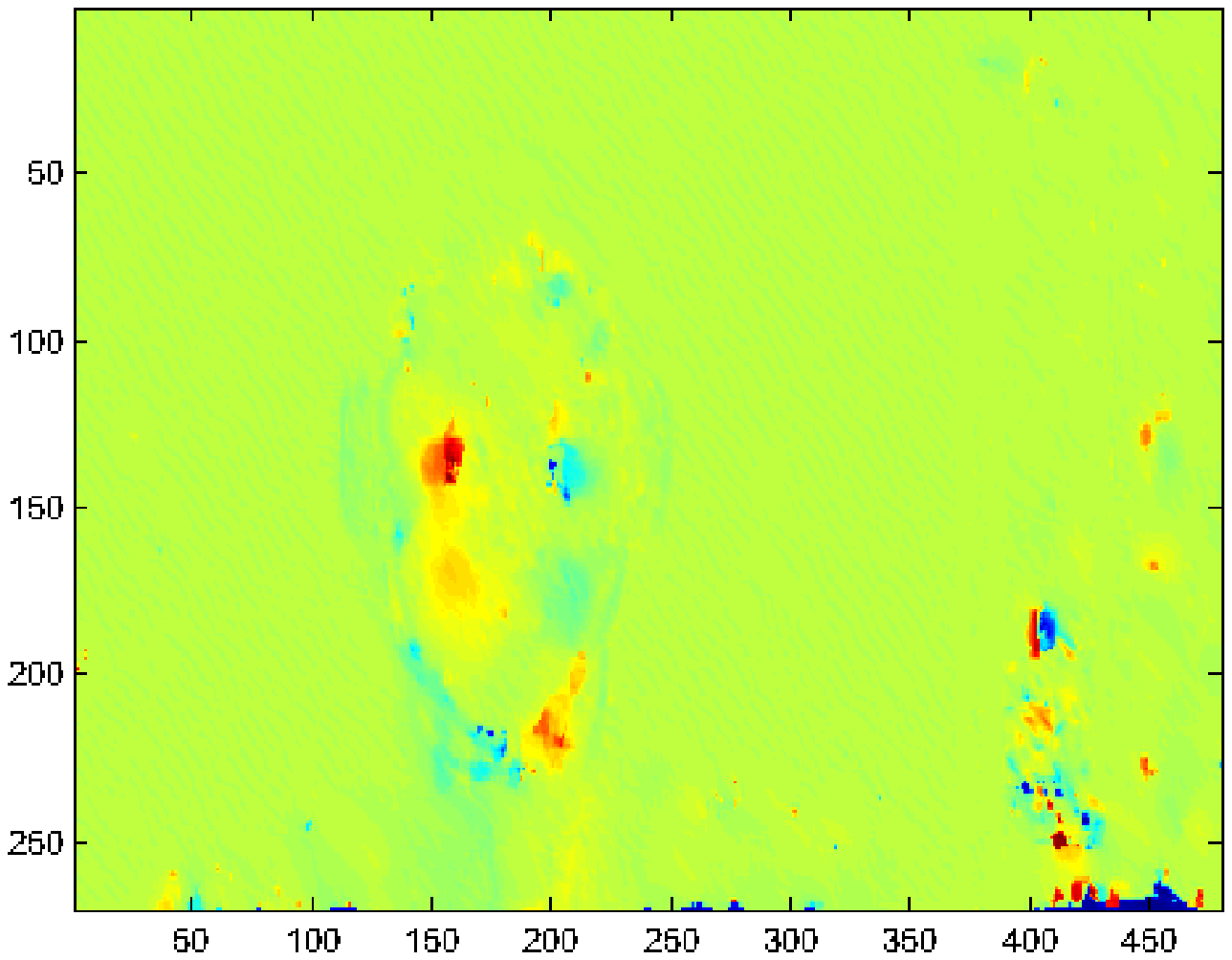}
\includegraphics*[height=1.8cm]{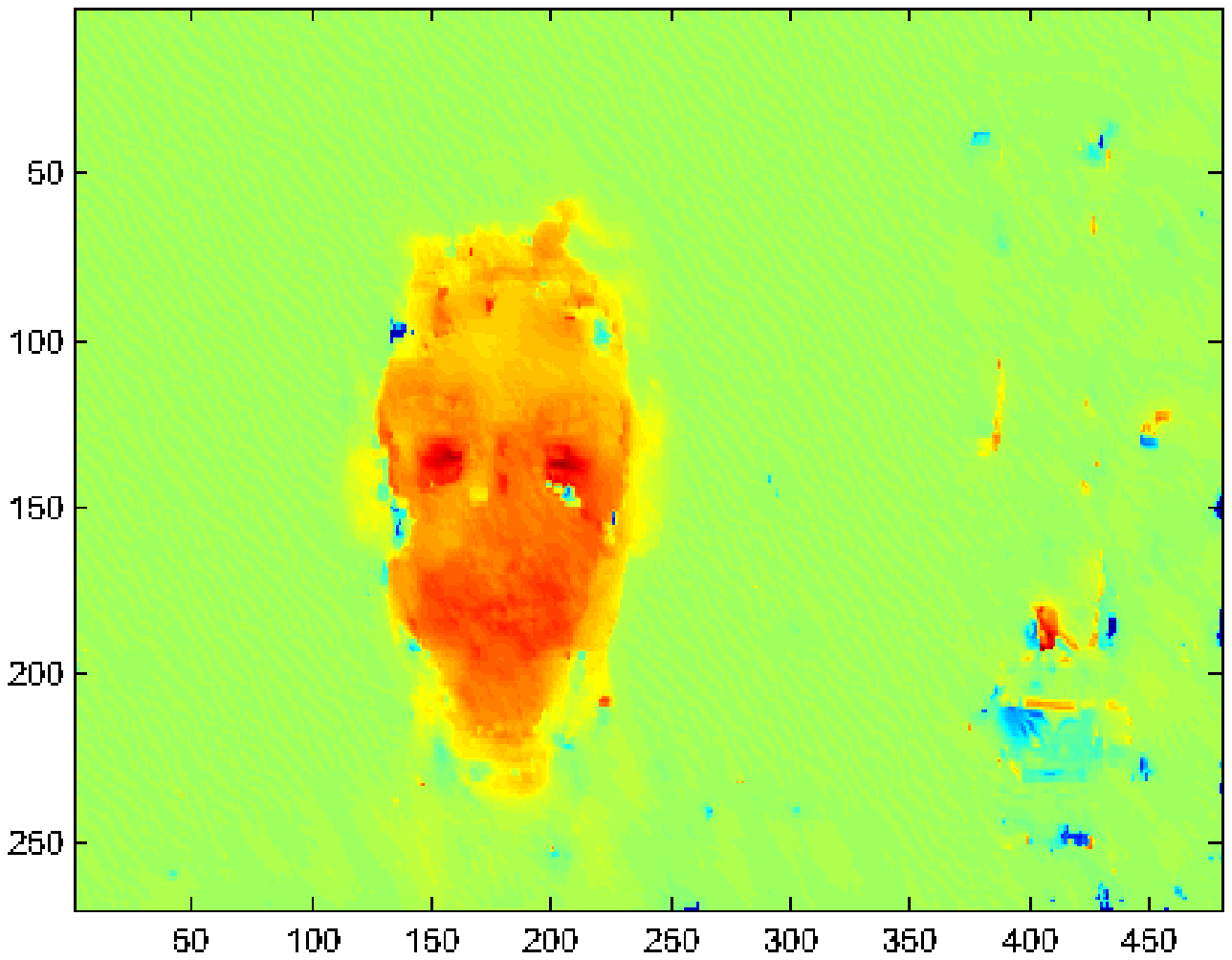}
\end{minipage}
\caption{Original images and their dense optical flows with their corresponding spontaneous and posed smiles variations of a subject. (Best viewed in color and zoomed in.)} \label{spontaneousPosedSmiles}
\end{figure}

In the third step, the optical flow is computed between the image at time $t$ and at time $t+1$ of the video sequence (see Fig. \ref{ImageandOpticalFlow}, top). The two components of the optical flow are illustrated in Fig. \ref{ImageandOpticalFlow}, bottom, which shows the optical flow along the x-axis and the optical flow along the y-axis. Because we are using a dense optic flow algorithm, the time to process one picture is relatively important. To speed up the processing, we computed the optic flow only in the three ROI regions: right eye, left eye and mouth. The optical flow computed in our approach is a pyramidal differential dense algorithm that is based on the following constraint: \vspace{-0.2cm}
\begin{equation}\label{eq:5}
F=F_{smooth}+\beta F_{attach},
\end{equation}
where the \textsl{attachment} term is based on thresholding method \cite{Zach07aduality} and the regularization term is based on the method developed by Meyer in \cite{Meyer1},  $\beta$  is a weight controlling the ratio between the end attachment and the term control. Ouarti \emph{et al.} in \cite{ouarti2013method} proposed to use a regularization that do not use an usual wavelet but a non-stationary wavelet packet \cite{ouarti2009method}, which generalize the concept of wavelet for extracting optical flow information. We extend this idea for extracting fine grained information for both micro and macro motion variations in smile videos as shown in Fig. \ref{fig3_leftEandrightE}. Fig. \ref{spontaneousPosedSmiles} shows the dense optical flows with spontaneous and posed smiles variations. \vspace{-0.5cm}
 \vspace{-0.5cm}
In the fourth step, for each of the three ROIs, the median of the optical flow is determined that give a cue to the global motion of the area. An histogram is also computed based on the optical flow that has 10 bins. The top three bins in term of cardinality are kept among all the bins. A linear regression is then applied to find the major axis of the point group for each of the three bins determined. In the end, for each ROI we obtain: the median value of the bin 1, the value of the bin 2 and the value of the bin 3. It also calculates the intercept and slope for points of bins 1, 2 and 3. These result in 60 features for each frame (using two consecutive frames in a smile video). SVM is then used on these features to classify the posed and spontaneous smiles. \vspace{-0.5cm}
 \vspace{-0.55cm}
The major advantage of this approach is that we can obtain useful smile discriminative features using a fully automatic analysis of videos, no marker are needed to be annotated by an operator/user. Moreover, rather than attempting to classify raw optical flow we design some processing to obtain a sparse representation of the optical flow signal. This representation helps in classification by extracting only the useful information in low dimensions and speeds up the calculation of the SVM. Finally, information is not completely connected to the positioning of the different ROI knowing that this positioning may vary from one frame to another, it is dependent on the depth and highly variable depending on the individuals. Therefore a treatment which would be too closely related to the choice of the ROI would lead to non-consistent results.

\section{Experimental Results}
We test our proposed algorithm on UvA-NEMO Smile Database \cite{Dibeklioglu1}, it is the largest and most extensive smile (both posed and spontaneous) database with videos from a total of 400 subjects, (185 female, 215 male) aged between 8 to 76 years old, giving us a total of 1240 individual videos. Each video consists of a short segment (3-8 seconds) of either posed or spontaneous smiles. The videos are extracted into frames at 50 frames per second. The extracted frames are also converted to gray scale and downsized to $480\times270$. In all the experiments, we split the database, in which $80\%$ is used as training samples and the remaining $20\%$ is used as testing samples. Binary classifier SVM in LIBSVM \cite{Hsu10apractical} is used to form a hyperplane based on the training samples. When a new testing sample is passed into the SVM it uses the hyperplane to determine which class the new sample falls under. This process is then repeated 5 times using a 5-fold cross validation method. To measure the subtle differences in the spontaneous and posed smiles we compute the confusion matrices between the two smiles so as to find out how much accuracy we can obtain in using each of them in the actual and classified separately. The results from all 5 processes are averaged and shown in confusion Tables 2-5 and compared with other methods in Table \ref{CCRComparison}.

\subsection{Results using parameters from the facial components}
Tables \ref{tab:EyeFeatures} and in bracket ($\cdot$) show the accuracy rates in distinguishing spontaneous smiles from the posed ones using eyes and lips features. The results show that the eye features play very crucial role in finding the posed smiles where as the lips features are important for spontaneous smiles. Overall we could obtain an accuracy of 71.14\% and 73.44\% using eyes and lips features respectively. Table \ref{tab:EyeLipFeatures} show the classification performance using combined features from eyes and lips. It is evident from the table that using these facial component features, pose smile can be classified better as compared to the spontaneous ones.
\begin{minipage}[b]{.5\textwidth} \footnotesize
  \centering
  \begin{tabular}{ | c | c | c |}
    \hline
    \backslashbox{Actual}{Classified} & Spontaneous & Posed\\ \hline
    Spontaneous & 60.1 (67.5) & 39.9 (32.5)\\ \hline
    Posed & 17.5 (20.4) & 82.5 (79.6)\\ \hline
  \end{tabular}
  \captionof{table}{The overall accuracy (\%) in classifying spontaneous and posed smiles using only the eyes features is 71.14\%. In bracket ($\cdot$) shows accuracy using only the lips features is 73.44\%.}
  \label{tab:EyeFeatures}
\end{minipage}\qquad
\begin{minipage}[b]{.5\textwidth} \footnotesize
  \centering
  \begin{tabular}{ | c | c | c |}
    \hline
    \backslashbox{Actual}{Classified} & Spontaneous & Posed\\ \hline
    Spontaneous & 65.3 & 34.7\\ \hline
    Posed & 16.3 & 83.7\\ \hline
  \end{tabular}
  \captionof{table}{The overall accuracy (\%) in classifying spontaneous and posed smiles using the combined features from eyes and lips is 74.68\%. (rows are gallery, columns are testing)}
  \label{tab:EyeLipFeatures}
\end{minipage}

\subsection{Results using Dense Optical flow}
We use the features using dense optical flow as described in Section \ref{SectionDenseOpticalFlow}, the movement in both X- and Y-directions are recorded between every consecutive frames of each video. The confusion matrices are shown in Tables \ref{tab:ClassicOpticalflowX}, in bracket ($\cdot$) and \ref{tab:ClassicOpticalflowXY}. It can be see from the tables that the performance of optical flow is lower as compared to the component based approach. However, the facial component based feature extraction method requires user initialization to find and track fiducial points, where the dense optical flow features are fully automatic. It does not require any user intervention, so it is more useful for practical applications like first-person-views (FPV) or egocentric views on wearable devices like Google Glass for improving real-time social interactions \cite{Mandal5,Gan2}.
\begin{minipage}[b]{.5\textwidth} \footnotesize
  \centering
  \begin{tabular}{ | c | c | c |}
    \hline
    \backslashbox{Actual}{Classified} & Spontaneous & Posed\\ \hline
    Spontaneous & 57.8 (58.3) & 42.2 (41.7)\\ \hline
    Posed & 39.8 (30.8) & 60.2 (69.2)\\ \hline
  \end{tabular}
  \captionof{table}{The accuracy (\%) in classifying spontaneous and posed smiles using our proposed X-directions dense optical flow is 59\%. In bracket ($\cdot$) the accuracy using our proposed Y-directions is 63.8\%.}
  \label{tab:ClassicOpticalflowX}
\end{minipage}\qquad
\begin{minipage}[b]{.5\textwidth} \footnotesize
  \centering
  \begin{tabular}{ | c | c | c |}
    \hline
    \backslashbox{Actual}{Classified} & Spontaneous & Posed\\ \hline
    Spontaneous & 58.0 & 42.0\\ \hline
    Posed & 45.1 & 54.9\\ \hline
  \end{tabular}
  \captionof{table}{The accuracy (\%) in classifying spontaneous and posed smiles using our proposed fully automatic system using X- and Y-directions of dense optical flow is 56.6\%.}
  \label{tab:ClassicOpticalflowXY}
\end{minipage}

\subsection{Results using both Component based features and Dense Optical Flow}
We combine all the features obtained from facial component based parameters and dense optical flow in to a single vector and apply SVM. Table \ref{AllFeatures} shows the confusion matrix using spontaneous and posed smiles. It can be seen that the performance of spontaneous smiles classification improved using features from dense optical flow. The experimental results in Table \ref{AllFeatures} show that both features from facial components and dense optical flows are important for improving the accuracy. Features from facial components (as shown in Table \label{tab:Table2Para}) are useful for encoding information arising from the muscle artifacts within a face, however, the regularized dense optical flow features helps in encoding fine grained information for both micro and macro motion variations in face smile videos. \vspace{-0.5cm}
\begin{table}
\centering \footnotesize
\begin{tabular}{|c|*{2}{c|}}
\hline
\backslashbox{Actual}{Classified} & Spontaneous & Posed\\
\hline
Spontaneous & 83.6 & 16.4\\
\hline
Posed & 22.9 & 77.1\\
\hline
\end{tabular}
\caption{The accuracy (\%) in classifying spontaneous and posed smiles using our proposed fused approach comprising of both features from facial components and dense optical flow is 80.4\%.} \label{AllFeatures}
\end{table}

\subsection{Comparison with Other Methods}
Correct classification rates using various methods on UvA-NEMO are shown in Table \ref{CCRComparison}. \vspace{-0.5cm}
\begin{table} \footnotesize
\centering
\begin{tabular}{|c|*{2}{c|}}
\hline
Method & Correct Classification Rate (\%)\\
\hline
Pfister \emph{et al.} \cite{Pfister1} & 73.1\\
\hline
Dibeklioglu \emph{et al.} \cite{Dibeklioglu2} & 71.1\\
\hline
Cohn \& Schmidt \cite{Schmidt1} & 77.3\\
\hline
Mid-level fusion \cite{Dibeklioglu1} & 87.0\\
\hline
Eyelid Features \cite{Dibeklioglu1} & 85.7\\
\hline
Ours Eye+Lips+dense optical flow & 80.4\\
\hline
\end{tabular}
\caption{Correct classification rates (\%) on UvA-NEMO database.}
\label{CCRComparison}
\end{table}

\vspace{-1.8cm}

\section{Conclusions}
Differentiating spontaneous smiles from the posed ones is a challenging problem as it involves extracting subtle minute facial features and learning them. In this work we have analysed features extracted from facial component based parameters using fuducial points markers and tracking them. We have also obtained fully automatic features from dense optical flow on both eyes and mouth patches. It has been shown that the facial component based parameters give higher accuracy as compared to dense optical flow features for smile classification. However, the former requires initialization of the fiducial markers on the first frame and hence, it is not fully automatic. Dense optical flow has advantage that the features can be obtained without any manual intervention. Combining the facial components parameters and dense optical flow gives us highest accuracy for classifying the spontaneous and posed smiles. Experimental results on the largest UvA-NEMO smile database shows the efficacy of our proposed method.

\bibliographystyle{splncs03}
\bibliography{BiblioFeb2015}

\end{document}